\documentclass{article}

    \PassOptionsToPackage{numbers, compress}{natbib}
 \usepackage[dblblindworkshop, final]{neurips_2025}
\workshoptitle{Frontiers in Probabilistic Inference: Sampling Meets Learning}

\usepackage[utf8]{inputenc} % allow utf-8 input
\usepackage[T1]{fontenc}    % use 8-bit T1 fonts
\usepackage{hyperref}       % hyperlinks
\usepackage{url}            % simple URL typesetting
\usepackage{booktabs}       % professional-quality tables
\usepackage{amsfonts}       % blackboard math symbols
\usepackage{nicefrac}       % compact symbols for 1/2, etc.
\usepackage{microtype}      % microtypography
\usepackage{xcolor}         % colors
\usepackage{graphicx}
\usepackage{multirow}
\usepackage{algorithm}
\usepackage{algorithmic}
\usepackage{amsmath}

\newtheorem{theorem}{Theorem}[section]

\newtheorem{lemma}[theorem]{Lemma}

\title{A Sampling-Based Domain Generalization Study \\ with Diffusion Generative Models}

\author{%
  \textbf{Ye Zhu\thanks{Work mainly completed when YZ was a postdoc at Princeton University.}~$^{1,2}$, Yu Wu$^{3}$, Duo Xu$^{4}$, Zhiwei Deng$^{5}$, Yan Yan$^{6}$, Olga Russakovsky$^{1}$}\\
  $^{1}$Department of Computer Science, Princeton Univeristy, USA\\
 $^{2}$LIX, École Polytechnique, IP Paris, France\\
 $^{3}$School of Computer Science, Wuhan University, China\\
  $^{4}$Canadian Institute for Theoretical Astrophysics (CITA), University of Toronto, Canada \\
  $^{5}$Google DeepMind, USA \\
  $^{6}$Department of Computer Science, University of Illinois Chicago, USA
  }
\begin{document}

\maketitle

\begin{abstract}
In this work, we investigate the domain generalization capabilities of diffusion models in the context of synthesizing images that are distinct from the training data. Instead of fine-tuning, we tackle this challenge from a sampling-based perspective using frozen, pre-trained diffusion models. Specifically, we demonstrate that arbitrary out-of-domain (OOD) images establish Gaussian priors in the latent spaces of a given model after inversion, and that these priors are separable from those of the original training domain. This OOD latent property allows us to synthesize new images of the target unseen domain by discovering qualified OOD latent encodings in the inverted noisy spaces, without altering the pre-trained models. Our cross-model and cross-domain experiments show that the proposed sampling-based method can expand the latent space and generate unseen images without impairing the generation quality of the original domain. We also showcase a practical application of our approach using astrophysical data, highlighting the potential of this generalization paradigm in data-sparse fields such as scientific exploration.
\end{abstract}

\section{Introduction}
\label{sec:intro}

Generalization ability, which enables a model to synthesize data from diverse domains, has long been a challenge for deep generative models. The current research trend focuses on leveraging larger models with more training data to facilitate improved generalization. The popularity of recent large-scale models, such as DALLE-2~\citep{ramesh2022dalle2}, Imagen~\citep{ho2022imagen}, and StableDiffusion~\citep{stablediff}, has demonstrated the impressive and promising representation capabilities of state-of-the-art (SOTA) diffusion generative models when trained on enormous image datasets. However, scaling up is not a panacea and does not fundamentally solve the generalization challenge. In other words, for data domains that remain sparse in these already giant datasets, such as astrophysical observation and simulation data, even SOTA models fail to synthesize data suitable for rigorous scientific research. In addition, scaling up requires extensive resources, severely limiting the number of research groups that are able to participate and contribute, and consequently hindering research progress. Given these concerns, our work focuses on studying generalization ability in a few-shot setup, where a pre-trained diffusion generative model and a small set of raw images different from its training domain are provided, with the ultimate objective of generating new data samples from the target OOD domain.

\begin{figure}[t]
    \centering
    \includegraphics[width=0.99\textwidth]{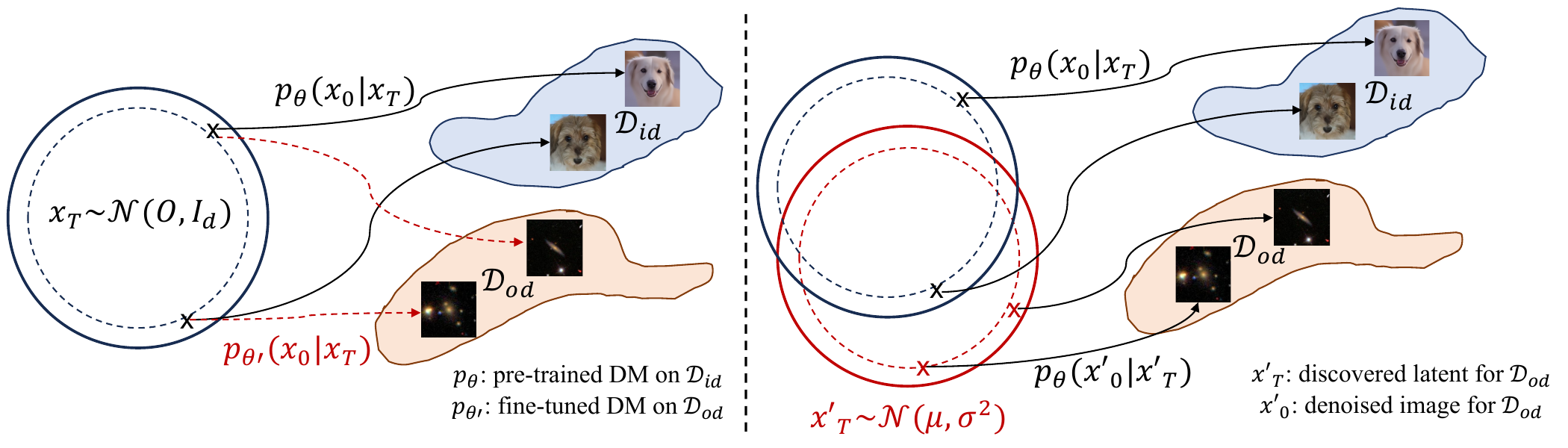}
    \caption{\textbf{Illustration of the trajectory-tuning based paradigm (left) and our proposed latent-sampling based paradigm (right) for OOD image synthesis with diffusion models.} Given a pre-trained DM $p_\theta$ on images from domain $\mathcal{D}_{id}$, most existing methods seek to finetune the generation trajectories $p_{\theta'}$ to synthesis data in a new domain $\mathcal{D}_{od}$. In contrast, we propose to discover unseen latent encodings to achieve the same goal via the frozen model $p_\theta$ by expanding the latent spaces.}
    \label{fig1:idea}
    \vspace{-0.12in}
\end{figure}

To achieve this objective, one of the most straightforward and intuitive approaches is to fine-tune the pre-trained model using the target OOD images~\citep{kim2022diffusionclip,zhao2022egsde,kwon2022diffusion}. However, tuning-based methods have several drawbacks, particularly when there is a large gap between the training and target data domains. It is well established that tuning methods \emph{do not generalize well} when the domain shift becomes too large~\citep{zhou2020learning,zhou2021domain,wang2019few}. In fact, our experiments show that fine-tuning pre-trained DMs using only image supervision with the vanilla variational lower bound loss~\citep{ho2020dpm} is \emph{very difficult}; performance improves only when additional semantic guidance, such as the CLIP loss~\citep{radford2021clip}, is introduced. Moreover, modifying model parameters can degrade synthesis quality on the original training domain, and the computational cost of tuning fully depends on the pre-trained model, which can be substantial given the well-known expense of training diffusion models.

In this work, we propose a sampling-based alternative to fine-tuning for tackling this problem. Our core insight lies in reframing the generalization challenge from “learning a new mapping function” to “discovering new OOD latents,” as illustrated in Fig.\ref{fig1:idea}. More specifically, we show that, after inversion, unseen OOD images exhibit several latent-space properties, including approximate Gaussianity and separability from the original training prior, as detailed in Sec.\ref{sec:probl}. The former allows for the discovery of new OOD latent encodings using relatively simple sampling techniques, while the latter ensures that new OOD samples can be generated without interference from the original generation trajectories. We validate the effectiveness of this approach through a series of experiments.

\vspace{-0.05in}
\section{Problem Formulation and Method}
\label{sec:probl}
\vspace{-0.05in}

\subsection{Problem Formulation}

Given a diffusion denoising probablistic model (DDPM) $p_\theta$ trained on images from a domain $\mathcal{D}_{id}$, we aim to investigate the generalization properties of $p_\theta$ on other domain $\mathcal{D}_{od}$ using $N$ data samples $\mathbf{x}_{od} \in \mathcal{D}_{od}$, and eventually generate new data samples $\mathbf{x}_{od}' \in \mathcal{D}_{od}$.
The objective of DDPMs is similar to most previous generative models, which is to approximate an implicit data distribution $q(\mathbf{x}_0)$ with a learned model distribution $p_\theta(\mathbf{x}_0)$, as well as providing an easy-to-sample proxy (e.g., standard Gaussian). 
% which is easy to sample from.
We further use $p_s$ and $p_i$ to represent the stochastic~\citep{ho2020dpm} and deterministic~\citep{song2020ddim} generation processes, respectively. 
For the opposite direction, the pre-defined diffusion procedure is often denoted by $q(\mathbf{x}_{1:T}|q_0)$.
% In addition, we use the hyper-parameter $\eta$~\citep{song2020ddim} to characterize the degree of stochasticity in the generative process, with $\eta = 1$ for $p_s$ and $\eta = 0$ for $p_i$. 
% At intermediate stochastic levels, we adopt the notation $p_{\eta=k}$ with $k$ equals a constant between 0 and 1. 
Similar to existing literature, $T$ denotes the total diffusion steps.
We use $\mathcal{X}_t$ to represent the latent (noisy) spaces formed by $\mathbf{x}_t$ along denoising.

% \vspace{-0.05in}
\subsection{Latent Sampling to New Domain Generalization}
% \vspace{-0.05in}

\begin{figure}[t]
    \centering
    \includegraphics[width=0.99\textwidth]{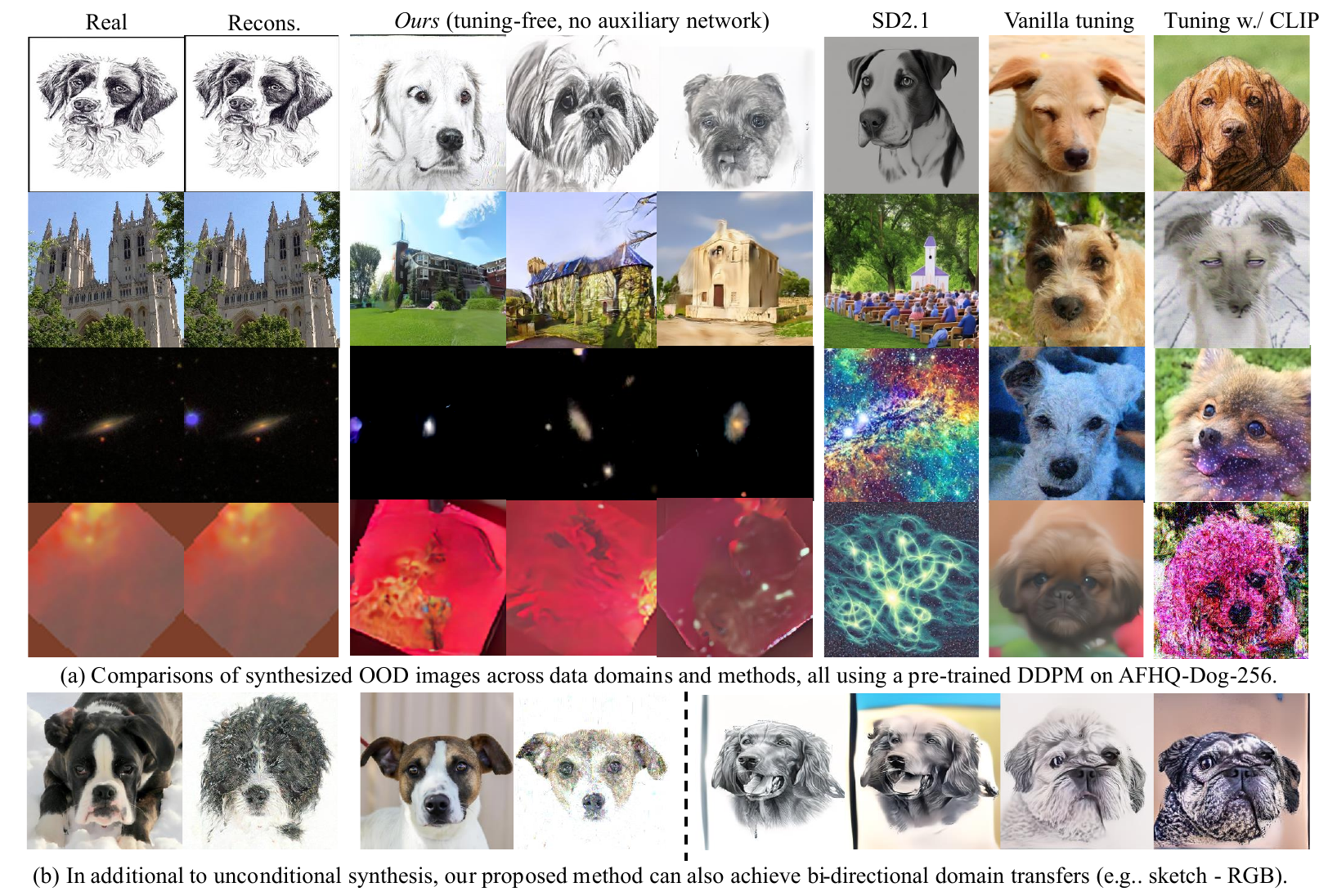}
    \caption{\textbf{Examples of synthesized OOD images across data domains and methods.} (a) All the OOD samples are obtained via our proposed sampling method using a pre-trained DMs on AFHQ-Dog~\citep{choi2020stargan}. (b) The same sampling method can also be applied to achieve style transfer (e.g., between RGB and sketch images).}
    
    % \emph{(Left):} Using a pre-trained DDPM on dog images as an example, we can well reconstruct arbitrary unseen images across domains covering natural images and astrophysical data. \emph{(Middle):} Even though large SOTA generative models have strong generalization abilities, their synthesized data are still very different from real data in specific domains. \emph{(Right):} In this work, we propose a sampling-based tuning-free method for synthesizing OOD data in a few-shot scenario. Different from tuning-based methods that usually yield better performance with closer domain gaps (e.g., dogs to humans), our method shows superiority as the domain gaps become larger.}
    \label{fig:teaser}
    % \vspace{-0.15in}
\end{figure}

\textbf{Latent Representation of Unseen Images.}
We observe that a DDPM, trained even on a single-domain small dataset (\textit{e.g.}, dog faces), already has sufficient representation ability to accurately reconstruct arbitrary unseen images (\textit{e.g.}, human, church, and astrophysical data), as shown in Fig.~\ref{fig:teaser}.
The reconstruction ability is subject to the deterministic inversion and denoising trajectories~\citep{song2020ddim}.
The findings above suggest that: with a good mapping approximator (i.e., pre-trained DDPM) and proper tool (i.e., deterministic trajectories with DDIMs), its intermediate latent spaces already have sufficient representation ability for arbitrary images, which opens up the possibility to leverage DDPMs for synthesizing images from new domains \emph{without tuning} the model parameters. Details about the deterministic inversion and reconstruction test can be found in the Appendix~\ref{sec:app_deterministic_diff}.

\textbf{Separability of Latent Gaussians.}
We then present two key properties of the latent OOD distributions after inversion, which are critical for developing an effective sampling-based method for domain generalization.
First, the inverted OOD latent encoding exhibits (approximate) Gaussians in the latent spaces. The Gaussianity is established with both theoretical and empirical groundings, with details about these proofs included in the appendix.
Second, the OOD priors are separable from the original Gaussian of the pre-trained image domains after the same inversion. We also show statistical support for this second property and its significance in the Appendix~\ref{sec:app_deterministic_diff}.

\textbf{Latent Sampling from OOD Priors.}
With the above properties, we then propose to generate new target images by first sampling from the OOD priors, and then denoise the newly sampled noisy latents via the deterministic DDIM denoising trajectories. Under the approximate Gaussian condition, the sampling is rather straightforward using the interpolation between arbitrary inverted OOD latents. Similarly, due to space constraints, we present additional details in the Appendix~\ref{sec:app_sampling}.

% \vspace{-0.1in}
\section{Experiments and Analysis}
\label{sec:exp_ans}
% \vspace{-0.1in}

\subsection{Experimental Setup}
\textbf{Model Zoos and Datasets.}
We adopt four pre-trained DDPMs on different single domain datasets as our base models for experiments: improved DDPM~\citep{nichol2021improveddmp} trained on AFHQ-Dog~\citep{choi2020stargan}, and DDPM~\citep{ho2020dpm} trained on CelebA-HQ~\citep{karras2017progressive}, LSUN-Church~\citep{yu2015lsun}, and LSUN-Bedroom~\citep{yu2015lsun}.
Each model generates images in the original resolution of $256\times256$, resulting in a total dimensionality of the latent spaces $d = 256 \times 256 \times 3 = 196,608$.

In addition to the above commonly used natural image datasets, we further experiment with two astrophysical datasets to cover a wide range of domain differences and to showcase the application scenarios with scientific data. 
Specifically, we adopt the GalaxyZoo~\citep{willett2013galaxy} and the radiation simulation data~\citep{xu2023predicting}, the latter has been investigated using DMs for prediction purposes. 
Details about those astrophysical datasets, their scientific interpretations, and evaluations are included in the Appendix~\ref{sec_appendix:experiments} for interested readers, which differs from the usual interpretation of natural images.

\textbf{Comparisons.}
Our main experiments focus on the comparison with the trajectory tuning-based approaches using the given DMs and OOD samples. 
Specially, we compare with vanilla tuning, where we finetune the pre-trained model with classic VLB loss~\citep{ho2020dpm} on unseen images, as well as several different SOTA image-to-image translation methods via diffusion models as baselines, as they also output images in domains that are different to the trained ones. However, it is worth noting that those baseline methods (\textit{i.e.}, DiffusionClip~\citep{kim2022diffusionclip}, and Asyrp~\citep{kwon2022diffusion}) are \emph{not really generating} unseen images, but rather \emph{editing an given original ID image} to a target unseen domain. Moreover, those are learning-based methods trained on each unseen domain with extra CLIP loss~\citep{radford2021clip}, while our method operates on a single mutual latent space from the pre-trained base diffusion models \emph{without} additional external supervision, echoing our conceptual design of ``space expansion''.

\textbf{Implementations and Learning Budget.}
We use $N=1000$ images for OOD domains, and the univariant Gaussian estimation for inverted OOD latent encodings.
% For deterministic diffusion, we adopt the standard DDIMs skipping step technique to accelerate both processes using 60 steps in total. 
% Each direction takes an average of 3 - 6 seconds.
% We use the univariant Gaussian estimation for inverted OOD latent encodings. 
% The geometric optimization tolerance $\omega$ is set to be 0.3 for distance for 0.1 for angles, which leads to a rejection rate of approximately 84.44 \% based on initial samples.
We use 2 RTX 3090 GPUs for all experiments including baselines.
For baseline methods that perform image translation and editing~\citep{kim2022diffusionclip,kwon2022diffusion}, we use their respective officially released implementations, each tuning process takes approximately 30 mins, with an initial default learning rate of 2e-6. 
To ensure a fair comparison, we also stop the vanilla tuning after around 30 minutes, resulting in around 20 epochs of finetuning, with an even larger learning rate of 1e-5.
For our proposed method, the inversion takes the same time as those baseline methods, but the core proposed sampling methods take \emph{negligible time}.

% \begin{figure}[t]
%     \centering
%     \includegraphics[width=0.99\textwidth]{./figs/teaser.pdf}
%     \caption{\textbf{Examples of synthesized OOD images across data domains and methods.} \emph{(Left):} Using a pre-trained DDPM on dog images as an example, we can well reconstruct arbitrary unseen images across domains covering natural images and astrophysical data. \emph{(Middle):} Even though large SOTA generative models have strong generalization abilities, their synthesized data are still very different from real data in specific domains. \emph{(Right):} In this work, we propose a discovery-based tuning-free paradigm for synthesizing OOD data in a few-shot scenario. Different from tuning-based methods that usually yield better performance with closer domain gaps (e.g., dogs to humans), our method shows superiority as the domain gaps become larger.}
%     \label{fig:teaser}
%     \vspace{-0.1in}
% \end{figure}

\textbf{Main Results.}
As the general quality evaluation, we calculate the FID scores~\citep{heusel2017fid} for natural images and report the Mean Opinion Scores (MOS) for astrophysical data.
The FID scores are averaged over four DDPMs pre-trained on different image domains, and the Mean Opinion Scores (MOS) with a scale between 1-5 are collected from subjective evaluations performed by astrophysicists with respect to the ground truth observation and simulation data in a \emph{non cherry-picky manner}. 
As shown in Fig.~\ref{fig:teaser} and Tab.~\ref{tab:main_result}, \emph{vanilla tuning} with only image supervisions can \emph{hardly} alter the original generation trajectories and synthesize desired images, always synthesizing in-domain images after comparable tuning time with other tuning baselines. 
As for methods that finetune the model with additional CLIP loss~\citep{radford2021clip}, such as DiffusionCLIP~\citep{kim2022diffusionclip} and Asyrp~\citep{kwon2022diffusion}, they relatively perform better for domains closer to their trained domains as expected.
Our proposed method shows an opposite trend by achieving better performance in data domains with bigger differences, as it is easier to avoid mode interference with larger domain gaps in the latent spaces.

\begin{table*}[t]
    \centering
    \caption{\textbf{General quality evaluation in cross model and domain setup.} \emph{T} stands for \emph{Tuning-based}, and \emph{TF} denotes \emph{Tuning-Free}. We report the FID scores ($\downarrow$) for natural image domains and the Mean Opinion Scores (MOS) ($\uparrow$) from subjective evaluations with astrophysicists. Note that most baseline methods perform the \emph{image-to-image translation} and use \emph{additional CLIP loss} to tune the model, and thus largely facilitate the task by bypassing the actual sampling stage and with extra strong semantic guidance. We therefore call for special attention when comparing the scores for a comprehensive and objective assessment, qualitative examples in Fig.~\ref{fig:teaser}.} 
    \label{tab:main_result}
    \scalebox{0.95}{
    \begin{tabular}{ccccccc}
    \toprule
       Methods   & Dog & CelebA & Church & Bedroom & Galaxy & Radiation  \\ \hline
        Vanilla tuning  & 213.6$\pm$4.8  & 229.7$\pm$4.3 &   192.5$\pm$3.7 & 191.1$\pm$4.0 & -& -\\
        CLIP tuning  & \textbf{73.6$\pm$2.9} & \textbf{63.6$\pm$3.0} & 66.3$\pm$2.8 & 68.1$\pm$2.8 & - & - \\ 
        % Asyrp~\citep{kwon2022diffusion} & \emph{T} & \textbf{66.4$\pm$2.1} & \textbf{58.7$\pm$2.4} & \textbf{62.5$\pm$2.0} & 63.2$\pm$1.9 & 1.84$\pm$0.97 & 1.35$\pm$0.74 \\ 
        Ours & 78.2$\pm$2.8 & 64.5$\pm$2.7 & \textbf{64.8$\pm$2.7} & \textbf{62.9$\pm$2.6} & \textbf{2.88$\pm$0.93} & \textbf{1.52$\pm$0.80} \\
        % Gaussian Sampling (Ours) & \emph{U} &  133.4$\pm$2.2  & 129.1$\pm$0.9 & 115.3$\pm$4.8 & 116.4$\pm$3.5 \\
        % Gaussian Sampling w/. Opt. (Ours) & \emph{U} & 117.7$\pm$2.6 & 114.4$\pm$4.8 &  103.8$\pm$8.0 &  105.6$\pm$7.4\\
        % Learning-Based Sampling (Ours) & \emph{U} & 80.6$\pm$1.2 & 80.9$\pm$1.4 & 77.1$\pm$1.3 & 79.9$\pm$0.7 \\

    \bottomrule
    \end{tabular}}
    % \vspace{-0.15in}
\end{table*}

\vspace{-0.05in}
\section{Conclusion}
\label{sec:conclu}
\vspace{-0.05in}

To sum up, we study the generalization abilities of diffusion models in the few-shot scenario. 
From the analytical point of view, we explore the generalization properties of diffusion models on unseen OOD domains.
From the methodological perspective, our analytical study allows us to propose a sampling-based method for synthesizing images from new domains without tuning the pre-trained generative trajectories.
In addition to experiments on natural images, we also showcase the superiority of our method in data-sparse cases with large domain gaps, such as in astrophysics.

\begin{ack}
This research was primarily conducted while YZ was a postdoctoral researcher at Princeton University. YZ also acknowledges travel funding from the French National Research Agency (ANR) via the “GraspGNNs” JCJC grant (ANR-24-CE23-3888), coordinated by Johannes F. Lutzeyer from École Polytechnique.
\end{ack}

% \newpage
{
    \small
    \bibliographystyle{ieeenat_fullname}
    \bibliography{main}
}

% \section*{References}

% References follow the acknowledgments in the camera-ready paper. Use unnumbered first-level heading for
% the references. Any choice of citation style is acceptable as long as you are
% consistent. It is permissible to reduce the font size to \verb+small+ (9 point)
% when listing the references.
% Note that the Reference section does not count towards the page limit.
% \medskip

% {
% \small

% [1] Alexander, J.A.\ \& Mozer, M.C.\ (1995) Template-based algorithms for
% connectionist rule extraction. In G.\ Tesauro, D.S.\ Touretzky and T.K.\ Leen
% (eds.), {\it Advances in Neural Information Processing Systems 7},
% pp.\ 609--616. Cambridge, MA: MIT Press.

% [2] Bower, J.M.\ \& Beeman, D.\ (1995) {\it The Book of GENESIS: Exploring
%   Realistic Neural Models with the GEneral NEural SImulation System.}  New York:
% TELOS/Springer--Verlag.

% [3] Hasselmo, M.E., Schnell, E.\ \& Barkai, E.\ (1995) Dynamics of learning and
% recall at excitatory recurrent synapses and cholinergic modulation in rat
% hippocampal region CA3. {\it Journal of Neuroscience} {\bf 15}(7):5249-5262.
% }

%%%%%%%%%%%%%%%%%%%%%%%%%%%%%%%%%%%%%%%%%%%%%%%%%%%%%%%%%%%%
\newpage
\appendix

\section{Technical Appendices and Supplementary Material}

We structure the appendices as follows:
We provide a detailed discussion on the related work in Appendix~\ref{sec:app_related}.
In Appendix~\ref{sec:app_deterministic_diff}, we present additional analysis of the latent OOD properties, including the background of deterministic diffusion models, the reconstruction test, the Gaussianity and the separability from the original ID latent prior.

\section{Related Work}
\label{sec:app_related}

% Due to the space limitations in the main paper, we present a detailed discussion of related work in our appendices.

\subsection{Generalization in Generative Models}
Domain Generalization~\citep{wang2022generalizing} that aims to generalize models' ability to extended data distributions has been an important research topic in broad machine learning area~\citep{ganin2016domain,zhao2020domain,zhou2021domain,muandet2013domain,li2017deeper}, with various computer vision applications such as recognition~\citep{peng2019moment,rebuffi2017learning}, detection~\citep{zhang2021understanding} and segmentation ~\citep{hoffman2018cycada,gong2019dlow}.
In the vision generative field, it becomes an even more challenging task with the extra demand to sample from the generalized distributions.
One popular recent trend in the computer vision community is scaling up the model and dataset sizes as the most intuitive and obvious solutions~\citep{ramesh2022dalle2,ho2022imagen,stablediff}.
Another scenario to study the domain generalization of generative models is within the few-shot scenario, where we only have a limited amount of data compared to the training set.
In this case, fine-tuning the given model on the limited images~\cite{kim2022diffusionclip} is the most straightforward way to go.

Our work falls into the second category: provided with a pre-trained model and a small set of unseen images different from the model's training domain, we seek to better understand the generalization abilities of DDPMs.

\subsection{Diffusion Models and Deterministic Variants}

Diffusion Models (DMs)~\cite{sohl2015dpm_thermo,ho2020dpm,song2019generative} are the state-of-the-art generative models for data synthesis in images~\citep{ramesh2022dalle2,stablediff,nichol2021improveddmp,gu2021vector,dhariwal2021diff-img1,hu2021global}, videos~\citep{ho2022video}, and audio~\citep{kong2020diffwave,zhu2022discrete,mittal2021symbolic-diff}. 
There are currently two mainstream fundamental formulations of diffusion models, i.e., the denoising diffusion probabilistic models (DDPMs)~\cite{ho2020dpm} and score-based models~\cite{theo-diff2}.
One common perspective to understand both formulations is to consider the data generation as solving stochastic differential equations (SDEs), which characterize a stochastic process.
Based on vanilla models, both branch develops their own deterministic variants, i.e., denoising diffusion implicit models (DDIMs)~\cite{song2020ddim} and consistency models~\cite{song2023consistency}, with their core idea to follow the marginal distributions in denoising.
Compared to initial DDPMs and Score-based DMs with ancestral sampling, the deterministic variants are solving ODEs instead of SDEs and largely accelerate the generation speed with fewer steps.

We leverage the deterministic variant (DDIMs~\cite{song2020ddim}) as the tool to achieve bidirectional transition between latent noisy space and data space in this work.

\subsection{Latent Space of Deep Generative Models}

Comprehensive studies of latent space of generative models~\citep{karras2017progressive,abdal2019image2stylegan,gal2022stylegan} help to better understand the model and also benefit downstream tasks such as data editing and manipulation~\citep{zhu2016generative,shen2020interpreting,kwon2022diffusion,zhu2020domain}. 
A large portion of work has been exploring this problem within the context of GAN inversion~\citep{xia2022gan}, where the typical methods can be mainly divided into either learning-based~\citep{zhu2016generative,richardson2021encoding,wei2022e2style,alaluf2021restyle} or optimization-based categories~\citep{abdal2019image2stylegan,abdal2020image2stylegan++,huh2020transforming,creswell2018inverting}.
More recently, with the growing popularity of diffusion models, researchers have also focused on the latent space understanding of DMs for better synthesis qualities or semantic control~\citep{stablediff,zhu2023boundary,yang2023cow}.

Our work also contributes to a better understanding of latent spaces, and aims to introduce a new synthesis paradigm to explore the intrinsic potential of DMs.

\subsection{Diffusion Models in Science}

While DMs have been extensively applied in data generation and editing within the multimodal context~\citep{stablediff,ho2022video,zhu2022discrete,yang2023cow,zhu2023boundary}, recent works have extended their application domains to scientific explorations, such as astrophysics~\citep{xu2023denoising,xu2023predicting}, medical imaging~\citep{kazerouni2023diffusion,wu2024medsegdiff}, and biology~\citep{wu2209protein}. Compared to conventional computer vision applications, scientific tasks usually exhibit several distinct features. For instance, data acquisition and annotation are generally more expensive due to their scientific nature, resulting in a relatively smaller amount of available data for experiments.
Additionally, the evaluation of these works adheres to established conventions within their respective contexts, which are usually different from image synthesis evaluation based on perceptual quality.

Our work also experiments with several astrophysical datasets to showcase the potential of applying our proposed paradigm and method to such specific domains with limited data.

\section{Additional Analysis of Latent OOD Properties}
\label{sec:app_deterministic_diff}

\subsection{Background of Deterministic Diffusion}

Our analytical studies and methodology designs are built upon a specific variant of diffusion formulations, i.e., the deterministic diffusion process.
While the original diffusion denoising probabilistic models (DDPMs) involve a stochastic process for data generation via denoising (\textit{i.e.}, the same latent encoding will output different denoised images every time after the same generative chain), there is a variant of diffusion model that allows us to perform the denoising process in a deterministic way, known as the Denoising Diffusion Implicit Models (DDIMs)~\citep{song2020ddim}.
DDIMs were initially proposed for the purpose of speeding up the denoising process, however, later research works extend DDIMs from faster sampling application to other usages including the inversion technique to convert a raw image to its arbitrary latent space in a deterministic and tractable way.
As briefly stated in our main paper, the core theoretical difference between DDIMs and DDPMs lies within the nature of forward process, which modifies a Markovian process to a non-Markovian one.

The key idea in the context of non-Markovian forward is to consider a family of $\mathcal{Q}$ of inference distributions, indexed by a real vector $\sigma \in \mathbb{R}^T_{\geq 0}$:
\begin{equation}
    q_{\sigma}(\mathbf{x}_{1:T}|\mathbf{x}_0) := q_{\sigma}(\mathbf{x}_T|\mathbf{x}_0)\prod^T_{t=2}q_\sigma(\mathbf{x}_{t-1}|\mathbf{x}_t,\mathbf{x}_0),
    \label{eq:9}
\end{equation}
where $q_\sigma(\mathbf{x}_T|\mathbf{x}_0) = \mathcal{N}(\sqrt{\alpha_T}\mathbf{x}_0, (1-\alpha_T)\mathbf{I})$ and for all $t > 1$,
\begin{equation}
    q_{\sigma}(\mathbf{x}_{t-1}|\mathbf{x}_t,\mathbf{x}_0) = \mathcal{N}(\sqrt{\alpha_{t-1}}\mathbf{x}_0 + \sqrt{1-\alpha_{t-1}-\sigma^2_t}\cdot\frac{\mathbf{x}_t-\sqrt{\alpha_t}\mathbf{x}_0}{\sqrt{1-\alpha_t}}, \sigma^2_t \mathbf{I}).
    \label{eq:10}
\end{equation}
The choice of mean function from Eqn.~\ref{eq:10} ensures that $q_\sigma(\mathbf{x}_t|\mathbf{x}_0) = \mathcal{N}(\sqrt{\alpha_t}\mathbf{x}_0,(1-\alpha_t)\mathbf{I})$ for all $t$, so that it defines a joint inference distribution that matches the ``marginals'' as desired.
The non-Markovian forward process can be derived from Bayes' rule:
\begin{equation}
    q_\sigma(\mathbf{x}_t|\mathbf{x}_{t-1},\mathbf{x}_0) = \frac{q_\sigma(\mathbf{x}_{t-1}|\mathbf{x}_t,\mathbf{x}_0)q_\sigma(\mathbf{x}_t|\mathbf{x}_0)}{q_{\sigma}(\mathbf{x}_{t-1}|\mathbf{x}_0)}.
    \label{eq:11}
\end{equation}

In fact, in the original paper, the authors also explicitly stated that: `` The forward process from Eqn.~\ref{eq:11} is also Gaussian (although we do not use this fact for the remainder of this paper)''.~\footnote{This paper refer to the DDIM paper~\citep{song2020ddim}.}
While this Gaussian property was not emphasized and leveraged in the DDIMs paper, we find it useful in our context to explore the representation and generalization ability of pre-trained DDPMs.

In particular, the hyper-parameters for Gaussian scheduling $\alpha$ and $\beta$ in the context of DDIMs are slightly different from the original formulation in DDPMs~\citep{ho2020dpm}. Denote the original sequences from DDPMs as $\alpha'_t$, then the $\alpha_t$ in this work follows the definition of DDIMs to be $\alpha_t = \prod^T_{t=1}\alpha'_t$.

In addition to DDIMs, we note that the score-based formulation has also recently marked a deterministic variant, namely the Consistency Models~\cite{song2023consistency}.
The core idea of the consistency model is, to some extent, similar to DDIMs, which allows the vanilla score-based stochastic diffusion models to achieve ``one-step'' denoising, by following the marginal distributions.

% To zoom out, the role of our

As mentioned in our main paper, the deterministic diffusion is mainly used as a tool in this work for our proposed tuning-free paradigm.

\begin{table}[t]
    \centering
    \caption{\textbf{Reconstruction results for arbitrary images via deterministic diffusion.} We use an iDDPM~\citep{nichol2021improveddmp} trained on AFHQ-Dog and 1K testing OOD images to compute the MAE (mean absolute error) reconstruction metric. Note DDIMs~\citep{song2020ddim} was initially proposed to accelerate DDPMs sampling, but have not been studied in this OOD reconstruction setting.}
    \label{tab:recons_result}
    \scalebox{1.0}{
    \begin{tabular}{ccc}
    \toprule
       Method   &  Recons. Domain &  MAE ($\downarrow$) \\ \hline
     pSp~\cite{richardson2021encoding} & CelebA (ID)  & 0.079  \\ 
     e4e~\cite{tov2021designing} & CelebA (ID)  & 0.092 \\
     ReStyle~\cite{alaluf2021restyle} & CelebA (ID)  & 0.089 \\
     HFGI~\cite{wang2022high} & CelebA (ID)  & 0.062 \\ \hline
     \multirow{6}{*}{DDIMs~\cite{song2020ddim}} & Dog (ID)  & 0.073 $\pm$ 6e-4 \\
     & CelebA (OOD)  & 0.073 $\pm$ 8e-4\\
     & Church (OOD)  & 0.074 $\pm$ 8e-4\\
     & Bedroom (OOD)  & 0.072 $\pm$ 7e-4\\
     & Galaxy (OOD) & 0.067 $\pm$ 1e-3\\
     & Radiation (OOD) & 0.077 $\pm$ 9e-4\\
    
    \bottomrule
    \end{tabular}}
\end{table}

\subsection{Latent Representation Ability via Reconstruction}

In our work, we evaluate the latent representation capability through reconstruction experiments and report the corresponding quantitative results in Tab.~\ref{tab:recons_result}.
These experiments serve as a sanity check under the assumption that, given plausible noisy latent encodings, the model should at least be able to faithfully reconstruct arbitrary target OOD images.

\subsection{Parameter-Independent Properties: Gaussian Priors}
\label{subsec:gaussian_prior}

We seek a theoretically grounded explanation to the generalization properties of pre-trained DDPMs after the inversion.
The takeaway message is: \emph{In theory}, the inverted latent encodings also establish Gaussian priors as presented in Lemma~\ref{lemma:latent_Gaussian}.\footnote{However, in practice, due to the fact that pre-trained DMs themselves are function approximators, the samples after inversion do not establish perfect Gaussians but rather approximations.}
\begin{lemma}
    For $q_\sigma(\mathbf{x}_{1:T}|\mathbf{x}_0$) defined in Eqn.~\ref{eq:9} and $q_\sigma(\mathbf{x}_{t-1}|\mathbf{x}_t,\mathbf{x}_0)$ defined in Eqn.~\ref{eq:10}, we have: 
    \begin{equation}
    q_\sigma(\mathbf{x}_t|\mathbf{x}_0) = \mathcal{N}(\sqrt{\alpha_t}\mathbf{x}_0,(1-\alpha_t)\mathbf{I}).
    \end{equation}
    \label{lemma:latent_Gaussian}
    \vspace{-0.2in}
\end{lemma}
As also mentioned in~\cite{song2020ddim}, one can derive Lemma~\ref{lemma:latent_Gaussian} by assuming for any $t \leq T$, $q_\sigma(\mathbf{x}_t|\mathbf{x}_0) = \mathcal{N}(\sqrt{\alpha_t}\mathbf{x}_0, (1-\alpha_t)\mathbf{I})$ holds, if:
\begin{equation}
    q_\sigma(\mathbf{x}_{t-1}|\mathbf{x}_0) = \mathcal{N}(\sqrt{\alpha_{t-1}}\mathbf{x}_0, (1-\alpha_{t-1})\mathbf{I}),
\end{equation}
and then prove the statement with an induction argument for $t$ from $T$ to 1, since the base case $(t=T)$ already holds by definition. We provide the detailed proof below.

\textit{Proof:}\\
Assume for any $t \leq T$, $q_\sigma(\mathbf{x}_t|\mathbf{x}_0) = \mathcal{N}(\sqrt{\alpha_t}\mathbf{x}_0, (1-\alpha_t)\mathbf{I})$ holds, if:
\begin{equation}
    q_\sigma(\mathbf{x}_{t-1}|\mathbf{x}_0) = \mathcal{N}(\sqrt{\alpha_{t-1}}\mathbf{x}_0, (1-\alpha_{t-1})\mathbf{I}),
\end{equation}
then we can prove that the statement with an induction argument for $t$ from $T$ to 1, since the base case $(t=T)$ already holds.

First, we have that 
\begin{equation}
    q_\sigma(\mathbf{x}_{t-1}|\mathbf{x}_0):= \int_{\mathbf{x}_t} q_\sigma(\mathbf{x}_t|\mathbf{x}_0)q_\sigma(\mathbf{x}_{t-1}|\mathbf{x}_t,\mathbf{x}_0)d\mathbf{x}_t,
    \label{eq:14}
\end{equation}

\begin{equation}
    q_\sigma(\mathbf{x}_t|\mathbf{x}_0) = \mathcal{N}(\sqrt{\alpha_t}\mathbf{x}_0,(1-\alpha_t)\mathbf{I}),
    \label{eq:25}
\end{equation}

\begin{equation}
    q_\sigma(\mathbf{x}_{t-1}|\mathbf{x}_t,\mathbf{x}_0) = \mathcal{N}(\sqrt{\alpha_{t-1}}\mathbf{x}_0+\sqrt{1-\alpha_{t-1}-\sigma^2_t}\cdot\frac{\mathbf{x}_t - \sqrt{\alpha_t}\mathbf{x}_0}{\sqrt{1-\alpha_t}},\sigma^2_t\mathbf{I}).
\end{equation}

According to~\cite{bishop2006pattern} \emph{2.3.3 Bayes' theorem for Gaussian variables}, we know that $q_\sigma(\mathbf{x}_{t-1}|\mathbf{x}_0)$ is also Gaussian, denoted as $\mathcal{N}(\mu_{t-1}, \Sigma_{t-1})$ where:
\begin{equation}
    \mu_{t-1} = \sqrt{\alpha_{t-1}}\mathbf{x}_0 + \sqrt{1-\alpha_{t-1}-\sigma^2_t} \cdot \frac{\sqrt{\alpha_t}\mathbf{x}_0 - \sqrt{\alpha_t}\mathbf{x}_0}{\sqrt{1-\alpha_t}} = \sqrt{\alpha_{t-1}}\mathbf{x}_0,
    \label{eq:17}
\end{equation}
\begin{equation}
    \Sigma_{t-1} = \sigma^2_t\mathbf{I} + \frac{1-\alpha_{t-1}-\sigma^2_t}{1-\alpha_t}(1-\alpha_t)\mathbf{I} = (1-\alpha_{t-1})\mathbf{I}.
\end{equation}
Therefore, $q_\sigma(\mathbf{x}_{t-1}|\mathbf{x}_0) = \mathcal{N}(\sqrt{\alpha_{t-1}}\mathbf{x}_0,(1-\alpha_{t-1})\mathbf{I})$, which allows to apply the induction argument.

\hfill \emph{Q.E.D}

\subsection{Data-Dependent Properties: Mode Interference and Separability}
\label{subsec:mode}

\begin{figure}[t]
    \centering
    \includegraphics[width=0.98\textwidth]{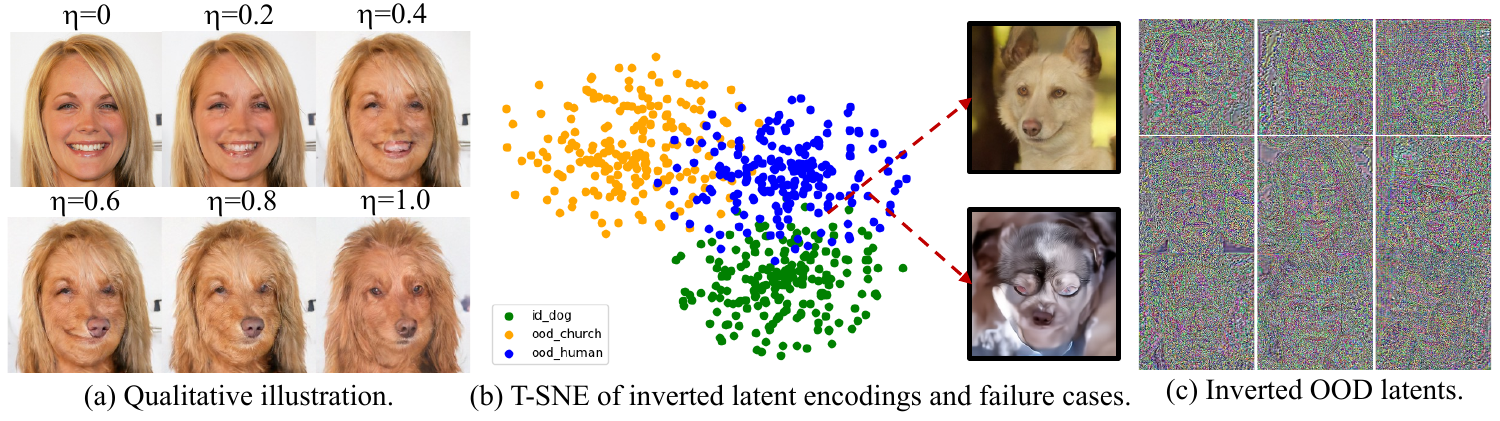}
    \caption{\textbf{Various visualizations of ``mode interference''.} Given an example setting of synthesizing human faces from DDPMs trained on dogs. \emph{(a):} An interfered image of human faces gradually becomes similar to its original trained domain as the denoising trajectory shifts from deterministic ($\eta$ = 0) to stochastic ($\eta$ = 1). \emph{(b):} Failure cases happen when sampled latent OOD encodings are captured by the model's original probabilistic concentration mass. \emph{(c):} Inverted OOD latent encodings preserve slight perceptible low-level visual features and are not \emph{perfect} Gaussians but rather approximations.}
    \label{fig:mode}
    % \vspace{-0.1in}
\end{figure}

In the literature of GANs-based generative models~\citep{gan}, ``mode collapse'' is a common issue that describes the training failure when generated images tend to be very similar given randomly sampled starting encodings from the Gaussian prior.
Within the context of diffusion models in our work, we explicitly reveal a phenomenon analog to the ``mode collapse'' in GANs, which we refer to as \emph{``mode interference''}, as qualitatively illustrated in Fig.~\ref{fig:mode} (a).

Intuitively, \emph{``mode interference''} describes the case when the denoised images fall into the model's original training domain $\mathcal{D}_{id}$ instead of the target unseen domain $\mathcal{D}_{od}$ due to the prior interference in the latent spaces. 
Specifically, when we sample directly from the standard Gaussian to obtain a latent encoding $\mathbf{x}_T \sim \mathcal{N}(\mathbf{0}, \mathbf{I}_d)$, then the denoised image will surely fall into the original training domain $\mathbf{x}_0 \in \mathcal{D}_{id}$ with $\mathbf{x}_0 \sim p_\theta(\mathbf{x}_0)$, which is the vanilla generation process of a trained DDPM.
However, it contradicts our task objective to synthesize images $\mathbf{x}'_0 \in \mathcal{D}_{od}$.
As illustrated in Fig.~\ref{fig1:idea}, since we are denoising the latent encoding via deterministic trajectories $p_i$, the remaining critical technical challenge to generate $\mathbf{x}'_0$ is to find additional qualified latent encoding $\mathbf{x}'_T$ \emph{free from the interference} of the ID Gaussian mode in the sampling stage.

Notably, a key precondition to achieving the effective OOD latent sampling is that the established OOD prior mode \emph{should be separable} from the ID Gaussian prior mode (i.e., a standard Gaussian). Otherwise, the denoised image would fall into the training domain as in Fig.~\ref{fig:mode} (b).
The separability is further supported and validated by our empirical verification below in Sec.~\ref{subsec:analytical_exp}.

\subsection{Analytical Experiments}
\label{subsec:analytical_exp}

We show empirical verification from multiple perspectives to support our parameter-independent and data-dependent properties described above.

\textbf{Geometrical Properties of Gaussians.}
We leverage the geometrical measurements established of the high-dimensional studies in mathematics~\citep{blum2020foundations}, as additional empirical support for the Gaussian priors in Sec.~\ref{subsec:gaussian_prior}.
Specifically, we compute several geometric metrics, including the pair-wise angles (angles formed by three arbitrary samples), sample-to-origin angles (angles formed by two arbitrary samples and the origin), pair-wise distance (euclidean distance between two arbitrary samples) and distance between OOD and ID Gaussian centers, and list the results in Tab.~\ref{tab:mode_study}.

\begin{table}[t]
    \centering
    \caption{\textbf{Geometric properties of inverted ID and OOD latent encodings.} The results are computed based on 1K sample pairs. We report the mean and std for each geometric measurement to ensure the statistical significance. The base model is trained on AFHQ-Dog~\citep{choi2020stargan} in 256x256. }
    \label{tab:mode_study}
    \scalebox{0.75}{
    \begin{tabular}{ccccccc}
    \toprule
        $\mathcal{D}$ & Dog (ID) & Human (O) & Bedroom (O) & Church (O) & Astro. Galaxy (O) & Astro. Turbulence (O) \\ \hline
        Pair-Angle & 60.0$\pm$0 & 60.0$\pm$0 & 60.0$\pm$0.1 & 60.0$\pm$0.1 & 60.0$\pm$0.1 & 60.0 $\pm$ 0\\
        Angle-Origin & 89.7$\pm$0.01 & 89.7$\pm$0 & 89.8$\pm$0.01 & 89.7$\pm$0.01& 89.1$\pm$0.01 & 87.6 $\pm$0.03 \\
        Pair-Distance & 607.4$\pm$0.01 & 611.4$\pm$0.05 & 611.2$\pm$0.07 & 609.9$\pm$0.02 & 612.37$\pm$0.05 & 609.13 $\pm$ 0.1\\
        Center-Distance & - & 33.3 &  26.0 & 33.2 & 54.7 & 60.8\\
        Clf. Acc. & - & 0.97 & 0.99 & 0.99 & 1.0 & 1.0 \\ 
    \bottomrule
    \end{tabular}}
    \vspace{-0.1in}
\end{table}

Characteristics above are typical geometric properties possessed by isotropic high-dimensional Gaussians~\citep{blum2020foundations}.
Notably, three randomly sampled points from a high-dimensional Gaussian are almost surely form an equilateral triangle and are almost surely nearly orthogonal, which corresponds to the constant 60$^\circ$ pair-wise angle and 90$^\circ$ sample-to-origin angle in the first and second rows of Tab.~\ref{tab:mode_study}, respectively.

\textbf{Mode Separability.}
% We also provide empirical verification for the separability of different Gaussian priors after inversion.
As revealed by our analysis in Sec.~\ref{subsec:mode}, the separability between ID and OOD Gaussian modes is critical for synthesizing target unseen domain images without modifying the model parameters and for avoiding the ``mode interference''. 
We further provide validation from the statistical and learning-based classifier perspectives to support the separability claim.

\textbf{Statistical Validation.}
The separability of high-dimensional Gaussians follows Lemma~\ref{lemma:gaussian_separability}~\citep{blum2020foundations}, which states that spherical Gaussians can be relaxingly separated by $\Omega(d^{\frac{1}{4}})$, or even $\Omega(1)$ with more sophisticated algorithms.
In other words, for a DDPM trained on 256$\times$ 256 images with dimensionality $d = 3 \times 256 \times 256$, ID and OOD modes can be well separated and avoid interference given a distance larger than $d^{\frac{1}{4}} \approx 21$, which is further validated by the empirical distance between centers, listed in the forth row of Tab.~\ref{tab:mode_study}.

\begin{lemma}
\label{lemma:gaussian_separability}
    Mixtures of spherical Gaussians in $d$ dimensions can be separated provided their centers are separated by more than $d^{\frac{1}{4}}$ distance (i.e., a separation of $\Omega(d^{\frac{1}{4}})$).
    and even by $\Omega(1)$ separation with more sophisticated algorithms.
\end{lemma}

\textit{Proof:}\\
% The detailed proof of this lemma depends on several other existing lemmas.

\begin{figure}[t]
    \centering
    \includegraphics[width=1.0\textwidth]{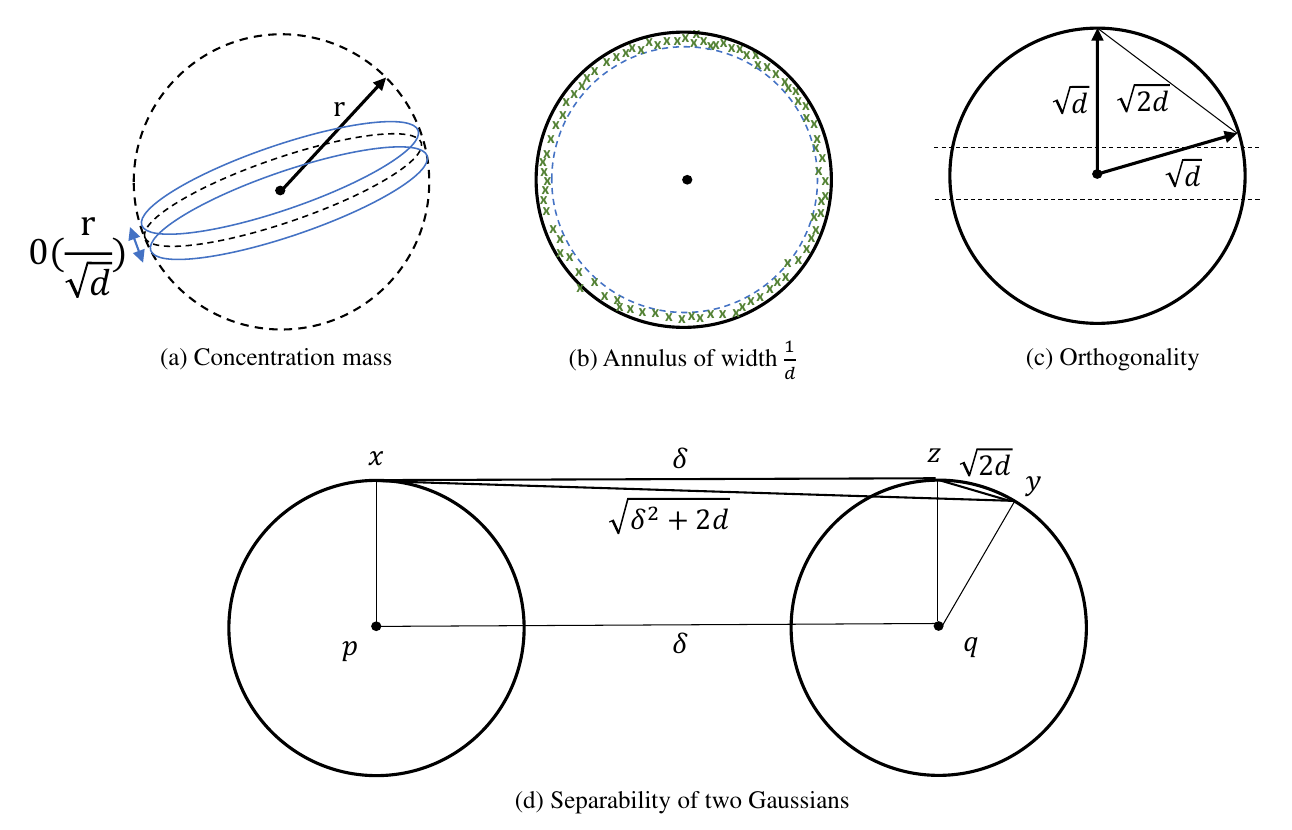}
    \caption{\textbf{Illustration of various geometric properties of high-dimensional Gaussians.} (a) and (b) show the probability concentration mass is mainly centered around a thin annulus around the equator. (c) illustrates the geometric observation on the orthogonality of sample pairs. (d) illustrates the idea of separating two Gaussian distributions in high-dimensional spaces. }
    \label{fig:geo}
\end{figure}

According to existing established understanding (Lemma 2.8 from ~\cite{blum2020foundations}), for a $d$-dimensional spherical Gaussian of variance 1, all but $\frac{4}{c^2}e^{-\frac{c^2}{4}}$ fraction of its mass is within the annulus $\sqrt{d-1} - c \leq r \leq \sqrt{d-1} + c$ for any $c > 0$, as illustrated in Fig.~\ref{fig:geo}. 

Given two spherical unit variance Gaussians, we have most of the probability mass of each Gaussian lies on an annulus of width $O(1)$ at radius $\sqrt{d-1}$. 
Also, $e^{-|x|^2/2}$ factors into $\prod_i e^{-x^2_i/2}$ and almost all of the mass is within the slab $\{\mathbf{x}|-c \leq x_1 \leq c\}$, for $c \in O(1)$.

Now consider picking arbitrary samples and their separability. After picking the first sample $\mathbf{x}$, we can rotate the coordination system to make the first axis point towards $\mathbf{x}$.
Next, independently pick a second point $\mathbf{y}$ also from the first Gaussian.
The fact that almost all of the mass of the Gaussian is within the slab $\{\mathbf{x}|-c \leq x_1 \leq c, c \in O(1)\}$ at the equator says that $\mathbf{y}$'s component along $\mathbf{x}$'s direction is $O(1)$ with high probability, which indicates $\mathbf{y}$ should be nearly perpendicular to $\mathbf{x}$, and thus we have $|\mathbf{x} - \mathbf{y}| \approx \sqrt{|\mathbf{x}|^2 + |\mathbf{y}|^2}$.

More precisely, we note $\mathbf{x}$ is at the North Pole after the coordination rotation with $\mathbf{x} = (\sqrt{(d)} \pm O(1), 0, ...)$. At the same time, $\mathbf{y}$ is almost on the equator, we can further rotate the coordinate system so that the component of $\mathbf{y}$ that is perpendicular to the axis of the North Pole is in the second coordinate, with $\mathbf{y} = (O(1), \sqrt{(d)} \pm O(1), ...)$.
Thus we have:
\begin{equation}
    (\mathbf{x} - \mathbf{y})^2 = d \pm O(\sqrt{d}) + d \pm O(\sqrt{d}) = 2d \pm O(\sqrt{d}),
\end{equation}
and $|\mathbf{x} - \mathbf{y}| = \sqrt{(2d)} \pm O(1)$.

Given two spherical unit variance Gaussians with centers $\mathbf{p}$ and $\mathbf{q}$ separated by a distance $\delta$, the distance between a randomly chosen point $\mathbf{x}$ from the first Gaussian and a randomly chosen point $\mathbf{y}$ from the second is close to $\sqrt{\delta^2+2d}$, since $\mathbf{x} - \mathbf{p}$, $\mathbf{p} - \mathbf{q}$, and $\mathbf{q} - \mathbf{y}$ are nearly mutually perpendicular, with:
\begin{equation}
    |\mathbf{x}-\mathbf{y}|^2 \approx \delta^2 + |\mathbf{z} - \mathbf{q}|^2 + |\mathbf{q} - \mathbf{y}|^2 = \delta^2 + 2d \pm O(\sqrt{d}).
\end{equation}

To ensure that the distance between two points picked from the same Gaussian are closer to each other than two points picked from different Gaussians requires that the upper limit of the distance between a pair of points from the same Gaussian is at most the lower limit of distance between points from different Gaussians. 
This requires the following criterion to be satisfied:
\begin{equation}
    \sqrt{2d} + O(1) \leq \sqrt{2d+\delta^2} - O(1),
\end{equation}
which holds when $\delta \in \Omega(d^{1/4})$.

Thus, mixtures of spherical Gaussians can be separated provided their centers are separated by more than $d^{1/4}$.

\hfill \emph{Q.E.D}

\textbf{Classifier Validation.}
Another empirical perspective to validate the separability between modes in the latent spaces is using the classifiers as in existing literature~\citep{shen2020interpreting,zhu2023boundary}.
Specifically, a linear classifier such as SVMs~\citep{hearst1998support} can be fitted to test the separability between ID and OOD encodings in the latent spaces.
In our analytical experiments, we fit SVMs on 1K inverted ID and OOD samples following the 7:3 training-testing ratio, and report the test accuracy in Tab.~\ref{tab:mode_study}. 
As additional clarification, the classification results are obtained with the test on the latent space $\mathcal{X}_T$. Our rationale behind the choice of $T$ corresponds to the recent findings of DMs~\citep{zhu2023boundary,yang2023cow}, which indicates that $\mathcal{X}_T$, as the departure latent space, has the largest probabilistic support for the trained domain.
In other words, if the latent ID and OOD modes can be separated in $\mathcal{X}_T$, they can be separated more easily in other $\mathcal{X}_t$, for $t = \{T-1, ..., t, ... 1\}$.

\subsection{Geometric Properties}

We consistently observe three geometric properties for the inverted OOD latent encodings.
We provide a more detailed discussion on what each property implies in this sub-section.

The three geometric properties as below:

\noindent \textit{\textbf{Observation 1:} For any OOD sample pairs $\mathbf{x}^{out}_{inv,i}$ and $\mathbf{x}^{out}_{inv,j}$ from the sample set, the Euclidean distance between these two points is approximately a constant $d_o$.}

\noindent \textit{\textbf{Observation 2:} For any three OOD samples $\mathbf{x}^{out}_{inv,i}$, $\mathbf{x}^{out}_{inv,j}$ and $\mathbf{x}^{out}_{inv,k}$ from the sample set, the angle formed between $\vec{\mathbf{x}^{out}_{inv,k} \mathbf{x}^{out}_{inv,i}}$ and $\vec{\mathbf{x}^{out}_{inv,k} \mathbf{x}^{out}_{inv,j}}$ is always around 60$^\circ$.}

\noindent \textit{\textbf{Observation 3:} For any OOD sample pairs $\mathbf{x}^{out}_{inv,i}$ and $\mathbf{x}^{out}_{inv,j}$ from the sample set, let $O$ denote the origin in the high-dimensional space, the angle formed between $\vec{O\mathbf{x}^{out}_{inv,i}}$ and $\vec{O\mathbf{x}^{out}_{inv,j}}$ is always around 90$^\circ$.}

For the first observation, when the sample pairs keep approximately the same distance, the direct implication is that those samples are likely to be drawn from some convex region in the high-dimensional space~\cite{wang2012geometric}.
One typical example is the spherical structure, where every data points exhibit an equal distance from the center.

The second geometric property suggests that the unknown samples could lie on a regular lattice near a low-dimensional manifold or sub-manifold, where the local geometry of the manifold is approximately Euclidean. 
However, a less evident implication is that for samples drawn from a high-dimensional Gaussian, this property also holds, as detailed in the next section~\ref{subsec:app_high_Gaussian}, and illustrated in Fig.~\ref{fig:geo}(c).

The third geometry property implies that the sample points might be isotropic in nature, who are rotationally symmetric around any point in the space. 
Therefore, any two points drawn from the distribution are equally likely to lie along any direction in the space.
This property is also observed for a high-dimensional Gaussian~\cite{blum2020foundations}, whose covariance matrix is proportional to the identity matrix.

We acknowledge that to deduce a distribution in high-dimensional space solely based on its geometric properties is very challenging, and there may exist other complex distributions that exhibit similar properties we have observed.
However, combined with our theoretical analysis and empirical observations, the OOD Gaussian assumption seems to hold well.
Explicitly, we find the above geometric properties do not hold for images $\mathbf{x}_0$ from the data space. 
For instance, the angle of samples to the origin is approximately 75$^\circ$ rather than 90$^\circ$.

\subsection{High-Dimensional Gaussian}
\label{subsec:app_high_Gaussian}

Gaussian in high-dimensional space establishes various characteristic behaviors that are not obvious and evident in low-dimensionality.
A better understanding of those unique geometric and probabilistic behaviors is critical to investigate the latent spaces of DDMs, since all the intermediate latent spaces along the denoising chain are Gaussian as demonstrated and proved in our previous sections.

We present below several properties of high-dimensional Gaussian from~\cite{blum2020foundations}, note those are known and established properties, we therefore omit the detailed proofs in this supplement, and ask readers to refer to the original book if interested.

\noindent
\textit{\textbf{Property D.1.} }
\textit{The volume of a high-dimensional sphere is essentially all contained in a thin slice at the equator and is simultaneously contained in a narrow annulus at the surface, with essentially no interior volume. Similarly, the surface area is essentially all at the equator.}

This property above is illustrated in Fig.~\ref{fig:geo}(a)(b), where the sampled ID encodings are presented in a narrow annulus.

\noindent 
\textit{\textbf{Lemma D.2.}}
\textit{
    For any $c>0$, the fraction of the volume of the hemisphere above the plane $x_1 = \frac{c}{\sqrt{d-1}}$ is less than $\frac{2}{c} e^{-\frac{c^2}{2}}$.
}

\noindent
\textit{\textbf{Lemma D.3.}}
\textit{For a d-dimensional spherical Gaussian of variance 1, all but $\frac{4}{c^2}e^{-c^2/4}$ fraction of its mass is within the annulus $\sqrt{d-1}-c \leq r \leq \sqrt{d-1} + c$ for any $c > 0$.}

The lemmas above imply that the volume range of the concentration mass above the equator is in the order of $O(\frac{r}{\sqrt{d}})$, also within an annulus of constant width and radius $\sqrt{d-1}$. 
In fact, the probability mass of the Gaussian as a function of $r$ is $g(r) = r^{d-1}e^{-r^2/2}$.
Intuitively, this states the fact that the samples from a high-dimensional Gaussian distribution are mainly located within a manifold, which matches our second geometric observation.

\noindent 
\textit{\textbf{Lemma D.4.}}
\textit{
The maximum likelihood spherical Gaussian for a set of samples is the one over center equal to the sample mean and standard deviation equal to the standard deviation of the sample.}

The above lemma is used as the theoretical justification for the proposed empirical search method in~\cite{zhu2023boundary}.
We also adopt the search method using the Gaussian radius for identifying the operational latent space along the denoising chain to perform the OOD sampling.

\noindent
\textit{\textbf{Property D.5.}}
\textit{Two randomly chosen points in high dimension are almost surely nearly orthogonal.}

The above property corresponds to the \emph{Observation 3}, where two inverted OOD samples consistently form a 90$^\circ$ angle at the origin.

\section{More Details about the Latent Sampling Methods}
\label{sec:app_sampling}

Intuitively, unlike sampling from a pure Gaussian, sampling from an approximate-Gaussian in non-trivial.
Alternatively, the general property of Gaussian distributions ensures that any convex combination of two samples lies within the same distribution. This motivates our choice of a simple yet effective approach to identify additional qualified $\mathbf{x}'_{ood,T} \in \mathcal{X}_{ood,T}$ using latent interpolation, a technique widely adopted in generative modeling for geometric and spatial analysis~\cite{song2020ddim,shen2020interpreting,zhu2023boundary,baumann2024continuous,preechakul2022diffusion}.
Specifically, as shown in Algo.~\ref{algo:main}, new latent samples $\mathbf{x}'_{od,T}$ are obtained by interpolating between two known latent encodings $\mathbf{x}^i_{od,T}$ and $\mathbf{x}^j_{od,T}$, both inverted from raw data.

 \begin{algorithm}[t]
    \caption{Latent sampling-centric approach for domain generalization}
    \label{algo:main}
    \begin{algorithmic}
        \STATE {\bfseries Input:} Arbitrary pre-trained DDPM $p_\theta$ in $\mathcal{D}_{id}$, $N$ images $\mathbf{x}_{od} \in \mathcal{D}_{od}$.
        \STATE{\bfseries Output:} images of the unseen target domain $\mathbf{x}_{od}' \in \mathcal{D}_{od}$ \\
       \textit{\textcolor{gray}{// Get the inverted latent encodings $\mathbf{x}_{od,T}$ }} \\
       \STATE Define $\{\tau_s\}^{S_{inv}}_{s=1}$ s.t. $\tau_1 = 0$, $\tau_{S_{inv}} = T$ \\
    \FOR{$i = 1, 2, ..., N $}
       \FOR{$s = 1, 2, ..., T-1 $}
    \STATE $\epsilon \leftarrow p(\mathbf{x}^i_{od,\tau_s}, \tau_s) $ 
    \STATE $\mathbf{x}^i_{od,\tau_{s+1}} = \sqrt{\alpha_{\tau_s}}\mathbf{x}^i_{od,\tau_s} + \sqrt{1 - \alpha_{\tau_s}} \epsilon$
   \ENDFOR 
   \ENDFOR \\
\textit{\textcolor{gray}{// Get new ood encodings $\mathbf{x}'_{od,T}$ and denoise via DDIMs}}\\
\STATE $\mathbf{x}'_{od,T} \leftarrow \text{Interp}(\mathbf{x}^i_{od,T},\mathbf{x}^j_{od,T}), \text{for} \: i \neq j, \: (i,j)\in \{1,2,...,N\}  $\\
    \STATE $\mathbf{x}'_{od} \leftarrow p_\theta(\mathbf{x}'_{od,T}, T)$

    \end{algorithmic}
    % \vspace{-0.1in}
\end{algorithm}

\section{More Details for Generative Experiments}
\label{sec_appendix:experiments}

\subsection{Background and Evaluation about the Astrophysical Data}
\label{subsec:astro}

\textbf{Galaxy Data.}
The images from the GalaxyZoo dataset~\cite{willett2013galaxy} are observation data of galaxies that belong to one of six categories - elliptical, clockwise spiral, anticlockwise spiral, edge-on, star/don't know, or merger.
The original data format of those galaxy images are also RGB images, thus ``somewhat'' similar to natural images, but they contain important morphological information to study the galaxies in astronomy.

The evaluation of the synthesized galaxy data is based on the expertise of astrophysicists if they could reliably classify the generated images into one of the known categories.

\textbf{Radiation Data.}
For the radiation data from~\cite{xu2023predicting}, the original format is physical quantity instead of RGB images, which correspond to the dust emission.

Dust is a significant component of the interstellar medium in our galaxy, composed of elements such as oxygen, carbon, iron, silicon, and magnesium. Most interstellar dust particles range in size from a few molecules to 0.1 mm (100 $\mu$m), similar to micrometeoroids. The interaction of dust particles with electromagnetic radiation depends on factors like their cross-section, the wavelength of the radiation, and the nature of the grain, including its refractive index and size. The radiation process for an individual grain is defined by its emissivity, which is influenced by the grain's efficiency factor and includes processes such as extinction, scattering, absorption, and polarization.

In RGB images of dust emission, different colors represent emissions at three wavelengths: blue for 4.5 $\mu$m, green for 24 $\mu$m, and red for 250 $\mu$m. The blue color typically indicates short-wavelength dust emission from point sources, such as young stars or young stellar objects. The green color represents mid-wavelength dust emission from warm and hot dust. The red color signifies long-wavelength dust emission from cold dust.

Warm/hot dust emission (green) is usually found around stars, which appear as blue-colored dots. Since warm dust often mixes with cold dust on the outer edges of bubble structures, the resulting color is often yellowish. Cold dust extends farther from the stars, giving the background or areas outside star clusters a red appearance. In the case of massive star clusters, stellar feedback, such as radiation and stellar winds, can blow away the surrounding gas and dust, creating black or blank areas.
Typically, RGB images show more extensive red emission with some orange/yellow emission, displaying filamentary and bubble structures, along with blue and/or white dotted point source emissions.

The above background is considered as part of the underlying evaluation criteria when performing subjective evaluation on the quality of generated radiation data.

\textbf{Qualitative evaluation from astrophysicists.}
As the evaluation of astrophysical data requires deep domain expertise, we collaborated with astrophysicists to subjectively assess the quality of the generated data using Mean Opinion Scores (MOS). Specifically, we provided 50 non-cherry-picked generated samples from our unseen domain generalization experiments alongside 50 raw data samples from true physical simulations.
For each generated sample across the two astrophysical datasets, a score ranging from 1 to 5 was assigned relative to the raw data, where 5 indicates the highest quality and 1 is the lowest.

Overall, the final MOS ratings for our generated galaxy data and radiation data are 2.88$\pm$0.93 and 1.52$\pm$0.80, respectively. In comparison, the same number of samples generated using the CLIP fine-tuning method received lower MOS ratings of 1.84$\pm$0.97 and 1.35$\pm$0.74 for the galaxy and radiation datasets, respectively.
These results highlight the relative superiority of our proposed method in generating higher-quality samples, particularly for domains with complex structures and significant domain gaps.

\subsection{More Experimental Results}
\label{sec:app_synthesis_exp}

We provide extended discussions in this section for the readers who are interested in more subtitle experimental details.

\begin{figure}[t]
    \centering
    \includegraphics[width=1.0\textwidth]{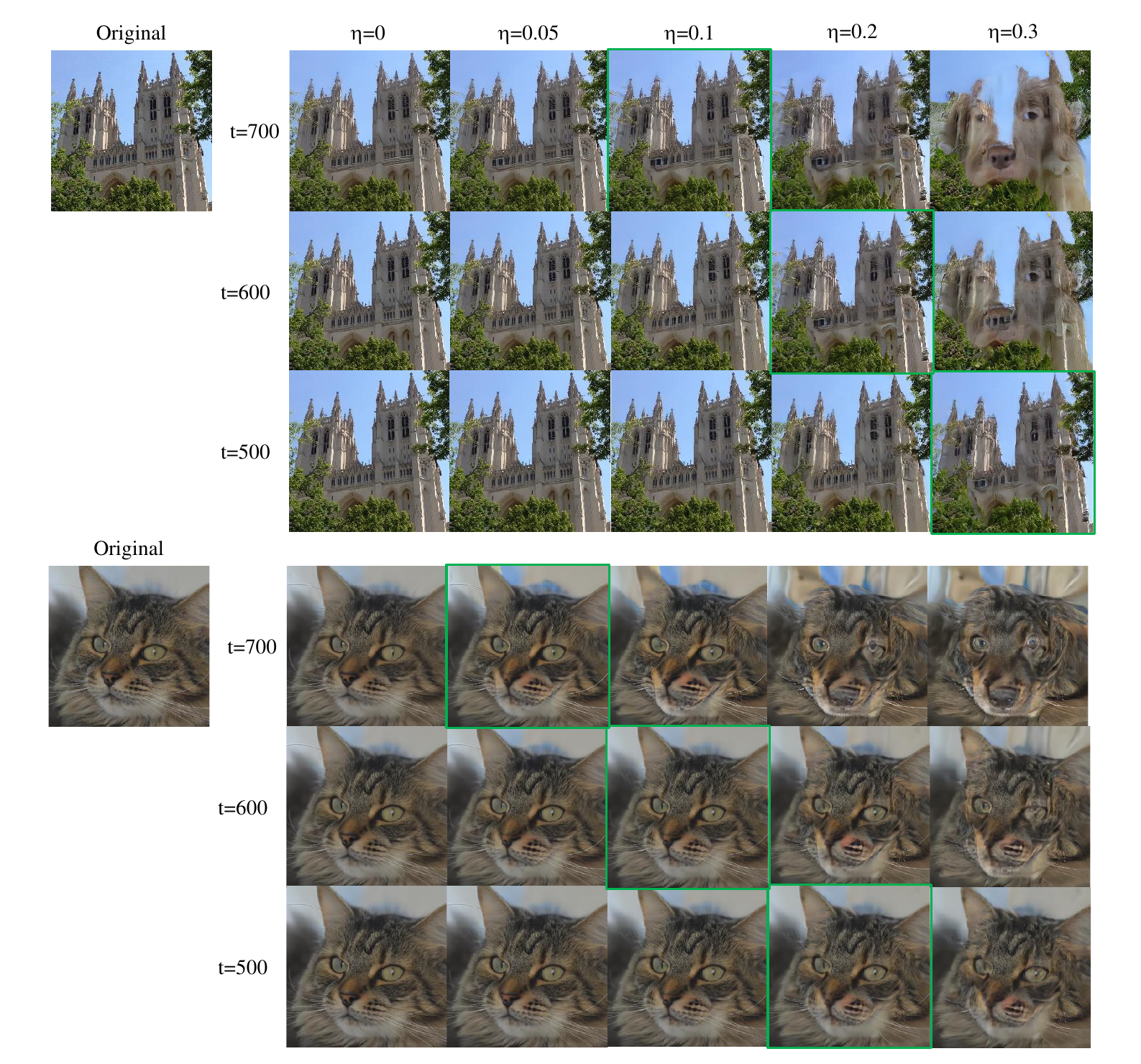}
    \caption{\textbf{Illustration of unseen trajectory bandwidth at different diffusion steps.}
    We show qualitative examples using the iDDPM~\cite{nichol2021improveddmp} trained on AFHQ-Dog-256 as the base model, the examples of church and cat are both unseen domain images. The image in green boxes indicates the bandwidth we have empirically selected to preserve the reconstruction quality. 
    Compared to the trained image domain (\textit{i.e.}, \textit{dogs}), \textit{cats} have a smaller domain gap than \textit{churches}. Different from the conventional understanding that a smaller domain gap is beneficial for better and easier generalization from a trained model, we observe a larger domain gap signifies a larger bandwidth, making it easier to perform the OOD sampling and synthesis.
    }
    \label{fig:sup_bandwidth}
\end{figure}

\subsubsection{Discussion on the Latent Step $t$, Stochasticity and Mode Interference}

% In our main paper, we briefly discuss the impact of the latent diffusion step $t$ where we perform the inversion and OOD latent sampling. 
While we empirically find that $t \approx 800$ is a reasonable range for the choice of $t$, we note there exists an entangled mechanism for the trade-off between the sampling difficulty and the mode interference issue.

For the diffusion step $t$, recent studies~\cite{zhu2023boundary,yang2023cow} suggest that $t$ characterizes the formation of image information at different stages of the denoising process. 
Intuitively, the early stage of the denoising process (e.g., $t > 800$) represents a rather chaotic process, the mixing step $t_{m}$~\cite{zhu2023boundary} signifies a critical stage where the image semantic information starts to form, and the later stage where $t$ is close to 0 demonstrates a stage during which more fine-grained pixel-level information are introduced to the final generated data.
From the distribution point of view, the influence of $t$ can be interpreted as the convergence of distributions, where $t=T$ is a standard Gaussian by definition, thus the ID and OOD modes are more difficult to separate.
However, as the denoising process gets closer to the real image space at $t=0$, the sampling difficulty increases as the implicit distribution moves away from the standard Gaussian.

Meanwhile, the diffusion step $t$ is not the only factor that impacts the trade-off between sampling difficulty and mode interference. 
While scarcely discussed in the main paper, we note the stochasticity of the denoising trajectory also plays a similar role as the diffusion step in this work.
The stochasticity of the denoising trajectory in DMs has been proven to be generally beneficial in improving the synthesis quality~\cite{karras2022elucidating,kim2022diffusionclip,kwon2022diffusion,zhu2023boundary}.
In this work, while we choose the $\eta = 0$ for the main paper, a tolerance for a certain range of stochasticity allows us to follow a ``relatively deterministic'' denoising process $p_{\eta =k}$, with $k \neq 0$, instead of the completely deterministic $p_i$. 
We hereby refer to it as ``bandwidth of the unseen trajectories,'' denoted as $\mathcal{B}_{\eta,t}$, which can be used to quantify the ``mode interference''.
Another interpretation is to analog the trajectory bandwidth $\mathcal{B}_{\eta,t}$ to the actual subspace volume occupied by the OOD latent samples. 
Fig.~\ref{fig:sup_bandwidth} shows more qualitative results for the bandwidth search in the reconstruction task and reveals its connection to the diffusion step $t$.
Overall, the bandwidth is a hyper-parameter that relates to the base model and the unseen domains, and the diffusion step $t$, while the bandwidth gets larger at the latent spaces closer to the raw image domains, sampling from OOD unseen distributions also gets more difficult.

\subsubsection{Discussion on Model Designs}

% We show the disentangled evaluations for different DDPM models, namely the improved DDPM~\cite{nichol2021improveddmp} and vanilla DDPM~\cite{ho2020dpm} in Tab.~\ref{tab:iddpm_dog}, Tab.~\ref{tab:ddpm_celeba}, Tab.~\ref{tab:ddpm_church} and Tab.~\ref{tab:ddpm_bedroom}, .
Among four base DDPMs we have tested, there are two architecture variants namely the improved DDPM~\cite{nichol2021improveddmp} and vanilla DDPM~\cite{ho2020dpm}. 
The difference between the two variants lies within the scheduler design for the Gaussian perturbation kernels: improved DDPM uses a cosine scheduler while vanilla DDPM adopts a linear one.
Our experiments suggest that iDDPM in general synthesizes images with better quality in terms of FID scores, which aligns with previous studies~\cite{nichol2021improveddmp,zhu2023boundary}.
One implication from the above observation is that the domain generalization abilities studied in this context is inherited from the performance of model's original performance.

\subsubsection{More Qualitative Results}

\begin{figure}[t]
    \centering
    \includegraphics[width=0.99\linewidth]{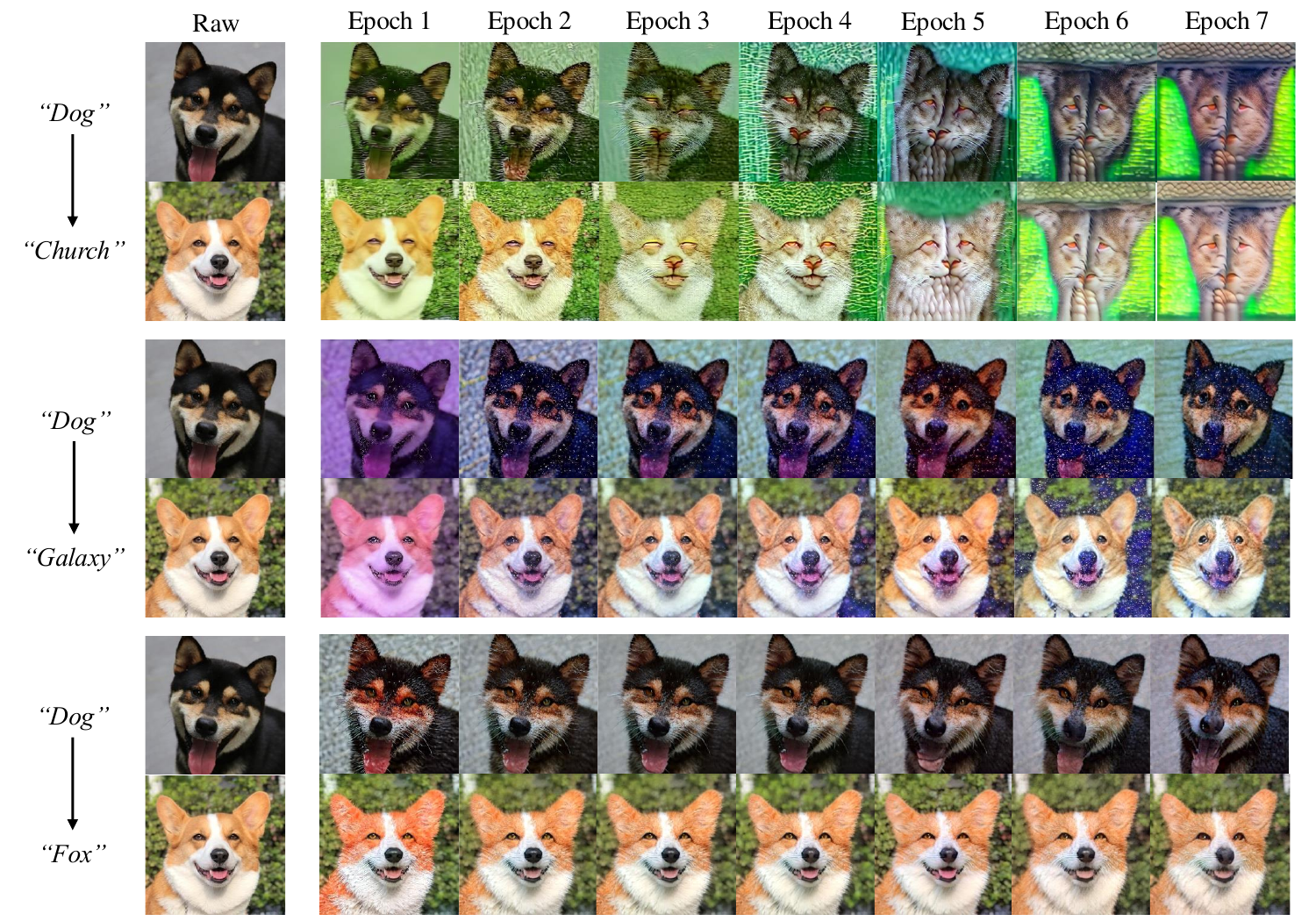}
    \caption{\textbf{Fine-tuning methods often fail to transfer the original trained domain to the target unseen domain with large distributional shifts.} We qualitatively show how a given ID sample (e.g., a dog RGB image) changes as the tuning epoch increases, using extra CLIP semantic guidance.
    }
    \label{fig:appendix_tuning_methods}
\end{figure}

We show more qualitative samples from the CLIP-tuned methods in Fig.~\ref{fig:appendix_tuning_methods} and note that the tuning-based methods often fail to generalize to new image domains with large distributional shofts.
% More synthesized examples of our proposed method are included in Fig.~\ref{fig:sup_results}.
% We also show part of the raw natural image samples used in our work in Fig.~\ref{fig:raw_human}, Fig.~\ref{fig:raw_church}, and Fig.~\ref{fig:raw_bedroom}, which helps to evaluate the diversity of the generated data.

\begin{figure}
    \centering
    \includegraphics[width=0.75\linewidth]{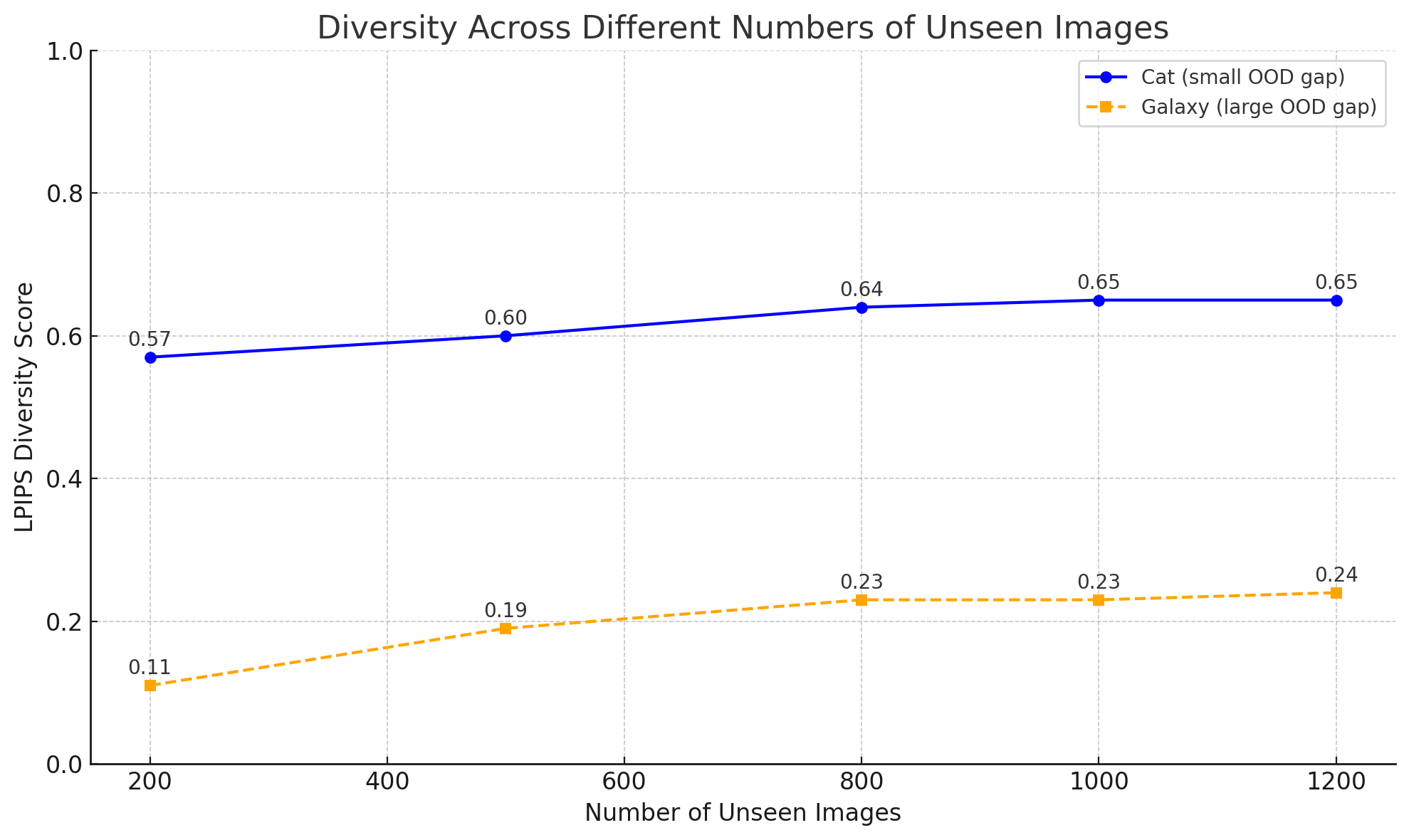}
    \caption{\textbf{Effect of the number of new samples from the target domains on generation diversity.} Results are shown for two OOD domains: Cat (small distribution gap) and Galaxy (large distribution gap) via diffusion model pre-trained on dogs.
}
    \label{fig:lipsp}
\end{figure}

In Fig.~\ref{fig:lipsp}, we first show the correlation between the diversity among generated samples with respect to the number of raw samples from the new target domains used in our proposed method. Specifically, we measure the diversity using the LPIPS scores, and note two takeaway information: First, the diversity increases when more samples are available, and our empirical findings suggest that $N = 800 - 1000$ is a reasonable choice. Next, the generated data from our proposed sampling method exhibits reasonable diversity among samples within the same target domain. Finally, it is important to note that the LPIPS score does not directly reflect the quality of the generated images. For example, in the case of galaxy images under large domain shift, the LPIPS score may appear lower despite the good perceptual quality of the outputs, since such images often share similar dark backgrounds and structural patterns by nature.

\begin{figure}[t]
    \centering
    \includegraphics[width=1.0\linewidth]{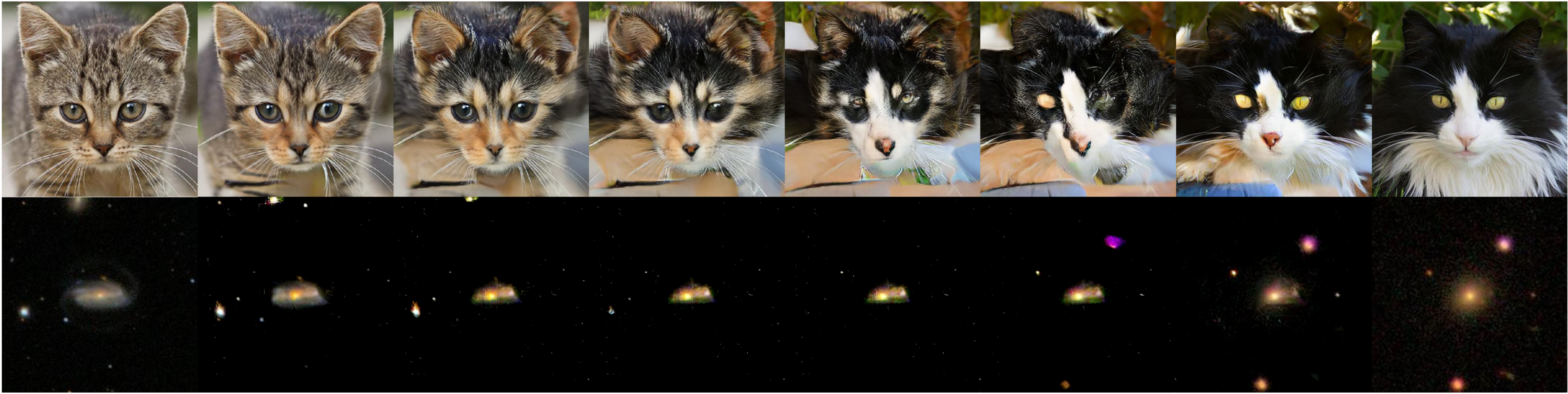}
    \caption{\textbf{Visualization of the interpolation between two raw samples.} Uncurated samples are shown for two OOD domains: Cat (small distribution gap) and Galaxy (large distribution gap) via diffusion model pre-trained on dogs.}
    \label{fig:interpolation}
\end{figure}

We also qualitatively visualize the generated data between the interpolation between two raw samples in Fig.~\ref{fig:interpolation}.
Overall, the results demonstrate noticeable visual variation across the full interpolation path, with higher similarity observed between samples that are spatially closer in the latent space. In practice, we typically select the samples from the middle of the interpolation, as it offers a better trade-off between quality and diversity.

\begin{figure}[t]
    \centering
    \includegraphics[width=0.9\linewidth]{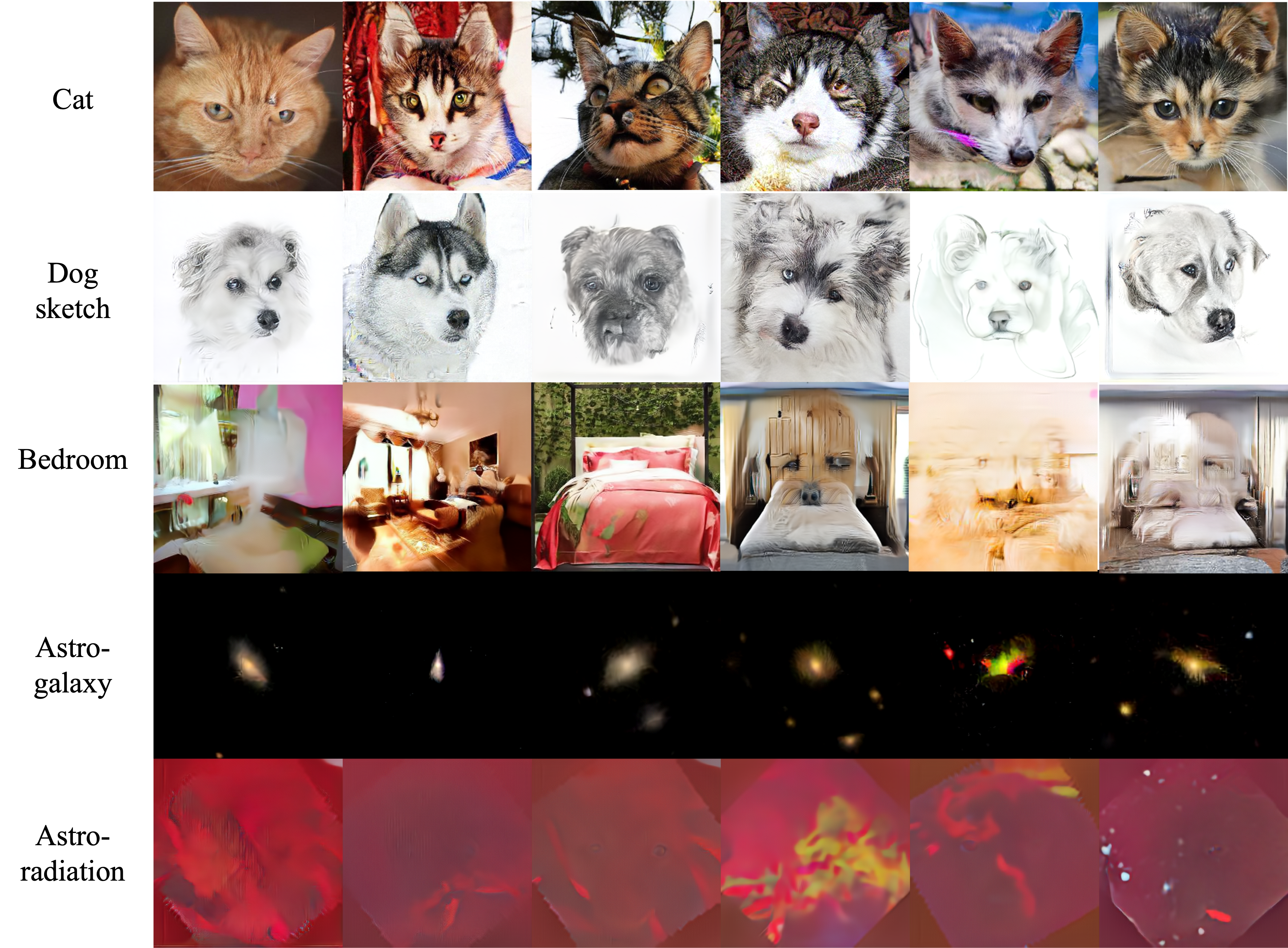}
    \caption{\textbf{More uncurated samples for various OOD domains generated via our proposed method from frozen diffusion models.}}
    \label{fig:uncurated}
\end{figure}

Finally, Fig.~\ref{fig:uncurated} includes more uncurated generation results from our proposed sampling-centric method from frozen pre-trained unconditional diffusion models.

\end{document}